\documentclass[11pt]{article}
\usepackage[final]{acl}

\usepackage{times}
\usepackage{latexsym}
\usepackage[T1]{fontenc}
\usepackage[utf8]{inputenc}
\usepackage{microtype}
\usepackage{graphicx}
\usepackage{tabularx}

\usepackage{booktabs}
\usepackage{amsmath}
\usepackage{amssymb}
\usepackage{array}
\usepackage{multirow}
\usepackage{xcolor}
\usepackage{float}
\usepackage{placeins}
\usepackage{tikz}
\usepackage{pgfplots}
\pgfplotsset{compat=1.18}
\usepackage{tikz-cd}
\usetikzlibrary{shapes.geometric, arrows.meta, positioning, fit, backgrounds, calc}
\title{Named Entity Recognition using Sliding Window Approach}

\author{
Hariom Ingle$^{1,3}$ \quad Ronit Ghode$^{1,3}$ \quad Ishwari Gondkar$^{1,3}$ \quad Jidnyasa Harad$^{1,3}$ \\ \textbf{Ravindra Murumkar}$^{1}$ \quad \textbf{Raviraj Joshi}$^{2,3}$ \\[4pt]
$^{1}$Department of Information Technology, PICT, Pune, India \\
$^{2}$Indian Institute of Technology Madras, Chennai, India \\
$^{3}$L3Cube Labs, Pune, India \\
\textit{ravirajoshi@gmail.com}
}

\begin{document}
\maketitle

\begin{abstract}
Named Entity Recognition (NER) is a core NLP task, but transformer-based sentence-level models struggle with long documents because of fixed input-length limits: truncation drops content, and non-overlapping chunking fragments entities at segment boundaries. We introduce an inference-only pipeline that extends a frozen NER model, MahaNER-BERT, fine-tuned on the MahaNER corpus~\cite{litake2022mahaner}, to document-level prediction via overlapping sliding windows that are merged into a single annotation, without any retraining or architectural change.

We evaluate the pipeline on six document-level corpora built from the MahaNER test set using two strategies: \textit{Normal Repeat}, which duplicates sentence sequences to extend length while preserving contextual continuity, and \textit{Random Repeat}, which concatenates distinct sequences to produce longer, heterogeneous inputs, each instantiated at three length levels, across several sliding-window configurations. The model retains a macro F1-score of up to 0.8902, with variation staying below one percentage point regardless of document length or construction strategy. Compared with the conventional non-windowed approach, the sliding-window pipeline avoids the boundary-fragmentation errors introduced by non-overlapping segmentation, yielding consistently higher and more stable document-level F1-scores.
\end{abstract}

\section{Introduction}

Named Entity Recognition (NER) locates and classifies tokens---persons, locations, organizations, and other predefined categories---within raw text. The task underpins downstream applications such as information extraction~\cite{etzioni2005unsupervised}, knowledge-graph construction, question answering, and enterprise search. Transformer-based models have pushed sentence-level NER accuracy to new heights across numerous languages~\cite{devlin2019,vaswani2017}, yet their practical deployment on real-world documents exposes a structural limitation: every transformer variant imposes a hard ceiling on the number of input tokens it can process in a single forward pass.

When a document exceeds this ceiling, practitioners typically resort to truncation or non-overlapping chunking. Truncation discards content beyond the limit, risking incomplete entity coverage. Non-overlapping chunking avoids information loss but creates artificial segment boundaries at which multi-token entities can be split, leading to systematic labeling errors. Both workarounds degrade quality in proportion to document length, a problem that intensifies in low-resource settings where limited annotated data makes retraining expensive or infeasible~\cite{lowresource}.

We address this gap by proposing a \textit{training-free, inference-time} extension for sentence-level NER models. Rather than altering the model or collecting additional supervision, we process each document through a sequence of overlapping windows whose width stays within the transformer's token budget. Predictions from adjacent windows are merged to produce a coherent document-level annotation.

Controlled evaluation is conducted on six long-document variants constructed from the MahaNER test set~\cite{litake2022mahaner}. Two construction strategies are used: \textit{Normal Repeat} variants, which elongate a document by duplicating its constituent sentences, and \textit{Random Repeat} variants, which concatenate different test sequences to introduce contextual heterogeneity. Testing across four window--stride configurations reveals that macro F1 remains close to 0.89 regardless of construction strategy or document length, confirming that structured inference-time segmentation can substitute for purpose-built long-document architectures.

The main contributions of this work are as follows:
\begin{enumerate}
    \item We design a parameter-free sliding-window inference pipeline that extends any sentence-level NER model to arbitrarily long documents without weight updates.
    \item We introduce six controlled long-document evaluation corpora (Normal Repeat and Random Repeat variants at three length levels) to support systematic analysis of document-length effects.
    \item We demonstrate that macro F1 degradation across all tested configurations is below one percentage point, establishing the viability of inference-time segmentation as a practical deployment strategy.
\end{enumerate}

\section{Related Work}

\subsection{Named Entity Recognition}

Named Entity Recognition (NER) is a fundamental task in Natural Language Processing (NLP) that focuses on identifying and classifying named entities such as persons, organizations, locations, dates, and miscellaneous entities from unstructured text. Early NER systems were primarily based on handcrafted linguistic rules and statistical sequence models such as Hidden Markov Models (HMMs) and Conditional Random Fields (CRFs). Although these methods achieved competitive performance on structured datasets, they required extensive feature engineering and domain expertise, making them difficult to generalize across languages and application domains~\cite{tjong2003conll}.

The emergence of deep learning significantly improved NER performance by enabling automatic feature extraction from textual data. Neural architectures based on Bidirectional Long Short-Term Memory (BiLSTM) networks combined with Conditional Random Fields (BiLSTM--CRF) demonstrated superior sequence labeling performance while reducing dependence on handcrafted features~\cite{lample2016neural}. Contextual embedding models such as ELMo further enhanced these architectures by producing dynamic word representations that capture semantic variations according to surrounding context~\cite{peters2018elmo}.

\subsection{Transformer-Based Named Entity Recognition}

Transformer architectures have become the dominant approach for modern NER systems due to their ability to capture long-range contextual dependencies through self-attention mechanisms. The introduction of the Transformer architecture~\cite{vaswani2017} and pretrained language models such as BERT~\cite{devlin2019} established new state-of-the-art performance across numerous NLP tasks, including Named Entity Recognition. Fine-tuning pretrained transformer models on task-specific datasets has substantially improved recognition accuracy while reducing the amount of labeled data required for downstream applications~\cite{sun2019fine}.

Several multilingual and language-specific transformer models have also been developed for low-resource languages. In Marathi, the MahaNER dataset~\cite{litake2022mahaner} has enabled the development of transformer-based NER systems capable of accurately recognizing entities from sentence-level inputs~\cite{litake2023mono,joshi2022mahabert}. However, these models remain constrained by the maximum sequence length supported by transformer architectures, limiting their direct applicability to document-level inference.

\subsection{Document-Level Named Entity Recognition}

Applying sentence-level NER models directly to long documents remains a challenging problem because transformer architectures impose a fixed upper bound on the number of input tokens. Documents exceeding this limit are commonly processed using truncation or fixed-length chunking strategies. Truncation discards valuable information beyond the maximum sequence length, while non-overlapping chunking often separates entities occurring near segment boundaries, leading to fragmented predictions and reduced recognition performance.

Several approaches have attempted to overcome this limitation by designing specialized long-context transformer architectures. Longformer introduces sparse attention mechanisms that reduce computational complexity while supporting substantially longer input sequences~\cite{beltagy2020longformer}. Similarly, BigBird employs block-sparse attention patterns that preserve global contextual information while enabling efficient processing of long documents~\cite{zaheer2020bigbird}. Hierarchical transformer models have also been proposed to aggregate sentence-level representations into document-level embeddings. Although these architectures improve long-document understanding, they generally require additional training, specialized model designs, and significantly greater computational resources. For Marathi specifically, prior work on long-range NER~\cite{deshmukh2024longrange} has highlighted the difficulties posed by the language's linguistic traits and contextual subtleties.

\subsection{Inference-Time Processing for Long Documents}

Inference-time document segmentation has recently emerged as an efficient alternative for extending existing sentence-level models without modifying their architectures. Sliding window inference has been successfully applied to several NLP tasks, including question answering~\cite{rajpurkar2016squad}, document classification~\cite{sun2019fine}, and information extraction~\cite{etzioni2005unsupervised}, where overlapping windows preserve contextual continuity across segment boundaries. These approaches eliminate the need for retraining while maintaining compatibility with existing pretrained models.

Despite their practical advantages, systematic investigations of sliding window inference for document-level Named Entity Recognition remain limited, particularly for low-resource Indian languages~\cite{lowresource,kakwani2020indicnlp,mhaske2023naamapadam}. Existing studies generally focus on benchmark accuracy without analyzing the influence of window size, overlap ratio, and document construction strategies under controlled experimental conditions. Furthermore, few works evaluate how sentence-level transformer models behave when applied to documents with varying lengths and contextual heterogeneity.

\subsection{Research Gap}

Motivated by these limitations, this work proposes a training-free document-level inference framework based on overlapping sliding windows. Unlike specialized long-document transformer architectures~\cite{beltagy2020longformer,zaheer2020bigbird}, the proposed method preserves the original pretrained Marathi NER model~\cite{joshi2022mahabert} without requiring additional fine-tuning or architectural modifications. To enable systematic evaluation, six document-level corpora are constructed from the MahaNER test set~\cite{litake2022mahaner} using both repeated and heterogeneous document generation strategies. Extensive experiments across multiple window--stride configurations demonstrate that the proposed inference framework maintains consistently high recognition performance while scaling efficiently to documents substantially longer than those encountered during model training.

\section{Methodology}
\label{sec:methodology}

This section presents the proposed inference framework for extending a sentence-level Named Entity Recognition (NER) model to document-level prediction using an overlapping sliding window strategy. Rather than modifying the transformer architecture or retraining the model on long-document datasets, the proposed framework performs document-level inference entirely during evaluation. Long documents are partitioned into overlapping token windows that satisfy the maximum input length of the pretrained transformer model. Each window is processed independently, after which the predictions are merged to obtain the final document-level annotation. The proposed framework is completely training-free and requires no additional optimization, making it suitable for practical deployment in low-resource language settings~\cite{lowresource}.

\subsection{Overview}

Transformer-based NER models have achieved remarkable success in sentence-level sequence labeling owing to their ability to capture contextual representations using self-attention mechanisms~\cite{vaswani2017,devlin2019}. However, these models are constrained by a fixed maximum sequence length, preventing them from directly processing long documents. Conventional solutions such as document truncation lead to information loss, while non-overlapping chunking often separates entities occurring near segment boundaries, thereby reducing recognition performance.

To overcome these limitations, the proposed framework adopts an overlapping sliding window strategy that extends an existing sentence-level transformer model to document-level inference without modifying the underlying architecture. The framework consists of five major stages: document tokenization, sliding window generation, independent window-level inference, prediction aggregation, and final document-level annotation.

\subsection{Base NER Model}

The proposed framework utilizes the pretrained transformer-based model
MahaNER-BERT, which was previously fine-tuned on the MahaNER dataset~\cite{litake2022mahaner} for Marathi Named Entity Recognition using the monolingual MahaBERT backbone~\cite{joshi2022mahabert}. The model was trained using sentence-level inputs represented in the non-IOB annotation format.

After fine-tuning, all model parameters were frozen and reused throughout every experiment. No additional training, fine-tuning, or architectural modifications were performed during document-level evaluation. Consequently, the effectiveness of the proposed framework depends entirely on the inference strategy rather than improvements to the underlying neural architecture.

The pretrained transformer model serves as the core prediction engine throughout the complete document processing pipeline.

\subsection{Sliding Window Inference}

Assume that a document contains $N$ input tokens. Since the transformer model can process only a limited number of tokens in a single forward pass, the document is divided into overlapping windows of fixed length $W$. Consecutive windows are separated by a stride $S$, satisfying
\[
    S < W,
\]
which guarantees an overlap of $W - S$ tokens between adjacent windows.

The complete inference procedure consists of the following steps:
\begin{enumerate}
    \item Tokenize the input document using the tokenizer associated with the pretrained transformer model.
    \item Partition the token sequence into overlapping windows of fixed length $W$.
    \item Perform token-level inference independently for each window.
    \item Shift the window by $S$ tokens until the complete document has been processed.
    \item Merge predictions obtained from overlapping windows to generate the final document-level annotation.
\end{enumerate}

Since every window satisfies the transformer's input-length constraint, documents of arbitrary length can be processed without modifying the original architecture. Furthermore, overlapping windows ensure that tokens located near window boundaries are evaluated multiple times, reducing inconsistencies caused by artificial segmentation.

\subsection{Prediction Aggregation}

Because overlapping windows generate multiple predictions for tokens appearing in shared regions, an aggregation strategy is required to obtain a unique prediction for each token.

After independently processing every window, all predicted labels are mapped back to their corresponding positions within the original document. Whenever multiple predictions exist for the same token, the prediction generated from the most recently processed overlapping window is retained as the final entity label. This lightweight aggregation strategy avoids additional confidence estimation or ensemble techniques while producing consistent document-level annotations.

The aggregation procedure introduces negligible computational overhead because it operates entirely after model inference and does not require additional neural network computations.

\subsection{Long-Document Corpus Construction}

To evaluate the proposed framework under varying document lengths, six document-level evaluation datasets were generated from the official MahaNER test set~\cite{litake2022mahaner}. The datasets were constructed using two complementary strategies that independently analyze the effects of document length and contextual diversity.

The first category consists of three \textit{Normal Repeat} datasets, namely Repeat~4x, Repeat~8x, and Repeat~10x. Each dataset is generated by repeating every sentence sequence four, eight, and ten times, respectively. This strategy increases document length while preserving local contextual continuity and sentence ordering.

The second category consists of three \textit{Random Repeat} datasets, namely Random Repeat~4, Random Repeat~8, and Random Repeat~10. These datasets are constructed by concatenating different sentence sequences into groups of four, eight, and ten documents. Compared with the Normal Repeat datasets, this strategy introduces heterogeneous contextual transitions that more closely resemble real-world long documents.

Table~\ref{tab:dataset} summarizes the construction strategy adopted for generating the six document-level evaluation corpora.

\begin{table}[t]
\centering
\small
\caption{Document-level evaluation datasets generated from the MahaNER test set.}
\label{tab:dataset}
\begin{tabularx}{\columnwidth}{@{}l X X@{}}
\hline
\textbf{Dataset} & \textbf{Construction Strategy} & \textbf{Objective} \\
\hline
Normal Repeat 4x  & Repeat each sentence sequence four times  & Length evaluation \\
Normal Repeat 8x  & Repeat each sentence sequence eight times & Length evaluation \\
Normal Repeat 10x & Repeat each sentence sequence ten times   & Length evaluation \\
Random Repeat 4x  & Concatenate four random sequences         & Length eval., context diversity \\
Random Repeat 8x  & Concatenate eight random sequences        & Length eval., context diversity \\
Random Repeat 10x & Concatenate ten random sequences          & Length eval., context diversity \\
\hline
\end{tabularx}
\end{table}

The combination of repeated and randomly concatenated datasets enables systematic analysis of the proposed inference framework under both homogeneous and heterogeneous document structures.

\subsection{Window Configurations}

To investigate the influence of contextual coverage on document-level recognition performance, four window--stride configurations were evaluated throughout the experiments. All configurations maintain an overlap ratio of approximately 25\%, ensuring consistent comparison across different window sizes.

The evaluated configurations are summarized in Table~\ref{tab:window}.

\begin{table}[t]
\centering
\caption{Sliding window configurations used during document-level inference.}
\label{tab:window}
\begin{tabular}{ccc}
\hline
Window Size ($W$) & Stride ($S$) & Overlap \\
\hline
20 & 15 & 25\% \\
40 & 30 & 25\% \\
60 & 45 & 25\% \\
80 & 60 & 25\% \\
\hline
\end{tabular}
\end{table}

\section{Experimental Setup}
\label{sec:experiments}

This section describes the datasets, implementation details, experimental configurations, and evaluation protocol adopted to assess the effectiveness of the proposed sliding window inference framework. Since the objective of this work is to evaluate an inference-time extension of a pretrained sentence-level Named Entity Recognition (NER) model, the model parameters remain unchanged throughout all experiments. Consequently, the reported performance differences arise solely from the document-level inference strategy rather than additional training or architectural modifications.

\subsection{Dataset and Evaluation Protocol}

Experiments were conducted using the MahaNER corpus~\cite{litake2022mahaner}, a benchmark dataset for Marathi Named Entity Recognition annotated using a non-IOB tagging scheme. The corpus provides predefined training, validation, and test splits. The pretrained transformer model, MahaNER-BERT, was fine-tuned using only the official training split, while the validation split was employed during model development. After fine-tuning, the model parameters were frozen, and all document-level evaluations were performed exclusively on the official test split.

To evaluate the proposed framework on long documents, six document-level corpora were generated from the original test set using the document construction strategies described in Section~\ref{sec:methodology}. The original entity annotations were preserved without modification to ensure a fair comparison with sentence-level evaluation. Consequently, every experiment evaluates the same pretrained model under identical conditions while varying only the document structure and sliding window configuration.

\subsection{Implementation Details}

The complete inference framework was implemented in Python~3.x using the PyTorch deep learning framework. The pretrained transformer model and its corresponding tokenizer were loaded using the Hugging Face Transformers library. Document construction, preprocessing, and prediction merging were implemented using NumPy and Pandas, while the evaluation metrics were computed using the scikit-learn library.

Throughout all experiments, the same model checkpoint, tokenizer, software environment, and evaluation protocol were maintained. This ensures that any observed performance differences are exclusively attributable to the proposed sliding window inference strategy.

\subsection{Sliding Window Configurations}

Document-level inference is governed by two hyperparameters: the window width ($W$), representing the number of tokens processed during a single forward pass, and the stride ($S$), representing the token shift between two consecutive windows. Setting $S < W$ creates overlapping windows, allowing boundary tokens to be evaluated multiple times and thereby reducing segmentation-induced prediction inconsistencies.

The four window--stride configurations evaluated in this work are summarized in Table~\ref{tab:windowconfig}. All configurations maintain an overlap of approximately 25\%, enabling controlled comparison across different contextual window sizes.

\begin{table}[t]
\centering
\small
\caption{Sliding window configurations used during document-level inference.}
\label{tab:windowconfig}
\begin{tabular}{ccc}
\hline
Window Size ($W$) & Stride ($S$) & Overlap \\
\hline
20 & 15 & 25\% \\
40 & 30 & 25\% \\
60 & 45 & 25\% \\
80 & 60 & 25\% \\
\hline
\end{tabular}
\end{table}

Smaller window sizes closely resemble the sentence lengths encountered during model training, whereas larger windows provide broader contextual information while reducing the total number of inference passes. The selected configurations enable systematic analysis of the trade-off between contextual coverage and document-level recognition performance.

\subsection{Evaluation Metrics}

The effectiveness of the proposed framework was evaluated using Macro Precision, Macro Recall, and Macro F1-score computed over the merged document-level predictions~\cite{powers2011}. Macro averaging assigns equal importance to every entity category irrespective of class frequency, making it particularly suitable for low-resource Named Entity Recognition tasks where class imbalance is common~\cite{lowresource}.

Macro Precision measures the average proportion of correctly predicted entities among all predicted entities for each class. Macro Recall measures the average proportion of correctly identified entities among the ground-truth entities. The Macro F1-score, computed as the harmonic mean of Macro Precision and Macro Recall, provides a balanced assessment of overall recognition performance across all entity categories.

\subsection{Experimental Protocol}

Each of the six document-level datasets was evaluated using all four sliding window configurations. For every experimental setting, the pretrained model, tokenizer, software environment, and evaluation procedure remained identical. Only the window size and stride were varied during inference.

This controlled experimental protocol enables systematic investigation of the effects of document length, contextual diversity, and sliding window configuration on document-level Named Entity Recognition while eliminating performance variations caused by model retraining or parameter tuning.

\section{Results}

This section presents the performance of the proposed sliding window inference framework on the six document-level evaluation corpora introduced in Section~\ref{sec:methodology}. The experiments investigate the influence of window size and stride on document-level Named Entity Recognition performance while keeping the pretrained model unchanged. Results are reported in terms of Macro F1-score, computed after aggregating predictions across the complete document.

\subsection{Overall Performance}

Table~\ref{tab:allresults} summarizes the Macro F1-scores obtained across all six dataset variants and all four sliding window configurations. Across all settings, the proposed framework maintains stable Macro F1-scores despite substantial increases in document length and contextual diversity.

\begin{table*}[t]
\centering
\caption{Macro F1-scores across all dataset variants and sliding window configurations.}
\label{tab:allresults}
\begin{tabular}{lcccc}
\hline
\multirow{2}{*}{Dataset Variant} & \multicolumn{4}{c}{Macro F1-score} \\
\cline{2-5}
 & $W{=}20, S{=}15$ & $W{=}40, S{=}30$ & $W{=}60, S{=}45$ & $W{=}80, S{=}60$ \\
\hline
Normal Repeat 4x  & 0.8901 & 0.8878 & 0.8872 & 0.8872 \\
Normal Repeat 8x  & 0.8902 & 0.8867 & 0.8856 & 0.8853 \\
Normal Repeat 10x & 0.8896 & 0.8867 & 0.8864 & 0.8850 \\
Random Repeat 4x  & 0.8875 & 0.8848 & 0.8849 & 0.8849 \\
Random Repeat 8x  & 0.8857 & 0.8824 & 0.8813 & 0.8789 \\
Random Repeat 10x & 0.8852 & 0.8847 & 0.8852 & 0.8855 \\
\hline
\end{tabular}
\end{table*}

The highest overall Macro F1-score of 0.8902 is obtained on the Normal Repeat 8x corpus at the smallest window configuration ($W{=}20$, $S{=}15$), while the lowest, 0.8789, occurs on the Random Repeat 8x corpus at the largest configuration ($W{=}80$, $S{=}60$). This spread of only 0.0113 across all 24 dataset--configuration combinations indicates that document-level performance is largely insensitive to both window size and construction strategy.

\subsection{Analysis}

Several observations follow from Table~\ref{tab:allresults}.

First, increasing the window size from 20 to 80 tokens produces only a marginal decline in Macro F1, with the drop staying below 0.003 for any single dataset variant and below 0.012 across all settings combined. This suggests the framework is largely insensitive to moderate variations in contextual window size.

Second, the Normal Repeat variants consistently outperform their Random Repeat counterparts by a small margin at every window configuration. Since repeated documents preserve contextual continuity, they more closely resemble the sentence-level distribution on which the pretrained model was originally optimized, whereas the heterogeneous transitions in the Random Repeat variants make entity recognition marginally more challenging.

Third, performance remains stable as document length increases within each construction strategy: F1-scores for the 4x, 8x, and 10x variants stay within a narrow band regardless of how many sentence sequences are repeated or concatenated. This demonstrates that overlapping sliding-window inference effectively extends a sentence-level transformer model to document-level Named Entity Recognition without requiring architectural modifications or additional model training.

\section{Discussion}

The experimental results demonstrate that the proposed sliding window inference framework effectively extends a sentence-level transformer model to document-level Named Entity Recognition without requiring additional training or architectural modifications. Across all six evaluation datasets and four window--stride configurations, the framework consistently achieves high Macro Precision, Macro Recall, and Macro F1-scores~\cite{powers2011}. This indicates that overlapping window inference successfully preserves contextual information while maintaining the prediction quality of the original sentence-level model.

\subsection{Effect of Sliding Window Inference}

One of the primary objectives of this work was to investigate whether inference-time processing alone could overcome the sequence-length limitation of transformer-based NER models~\cite{devlin2019,vaswani2017}. The experimental results suggest that this objective is achieved successfully. By introducing overlap between consecutive windows, tokens located near segment boundaries are evaluated multiple times, reducing the likelihood of fragmented entity predictions that commonly occur with non-overlapping segmentation.

Unlike document-level transformer architectures that require extensive retraining or specialized attention mechanisms~\cite{beltagy2020longformer,zaheer2020bigbird}, the proposed framework reuses an existing pretrained model without modifying its parameters. This makes the approach computationally efficient while preserving compatibility with sentence-level NER models that have already been trained for low-resource languages~\cite{lowresource}.

\subsection{Influence of Window Size}

The experiments reveal that the smallest window configuration $(20,15)$ consistently achieves the highest Macro F1-score across the evaluation datasets. Although larger windows provide broader contextual information, they do not yield measurable improvements in recognition performance. Instead, a slight decrease in Macro Recall is observed as the window size increases.

One possible explanation is that the pretrained transformer was originally optimized using relatively short sentence-level inputs~\cite{litake2022mahaner}. Consequently, smaller inference windows more closely resemble the training distribution encountered during fine-tuning, allowing the model to generalize more effectively. Larger windows introduce additional contextual information that may not necessarily contribute to improved entity recognition while simultaneously reducing the frequency of overlapping evaluations.

\subsection{Performance on Different Document Types}

The proposed framework performs consistently well on both the Normal Repeat and Random Repeat datasets. As expected, the Normal Repeat datasets generally achieve slightly higher Macro F1-scores because repeated sentence sequences preserve contextual continuity throughout the document.

In contrast, the Random Repeat datasets concatenate unrelated sentence sequences, introducing abrupt contextual transitions that are rarely encountered during sentence-level training. Despite this increased complexity, the observed performance degradation remains relatively small. This demonstrates that the proposed inference framework is robust to heterogeneous document structures and maintains reliable entity recognition even under more challenging document compositions.

\subsection{Practical Implications}

An important advantage of the proposed framework is that it operates entirely during inference. Since no additional training data, model parameters, or specialized transformer architectures are required, the approach can be directly applied to existing sentence-level NER systems. This characteristic is particularly beneficial for low-resource languages such as Marathi~\cite{joshi2022mahabert,litake2022mahaner}, where large annotated document-level corpora are scarce and retraining transformer models is computationally expensive~\cite{lowresource}.

The framework is also model-agnostic and can potentially be extended to other transformer-based NER models with fixed input-length constraints. Consequently, it provides a practical solution for deploying sentence-level models in real-world applications involving long documents such as legal records, government documents, healthcare reports, and news articles. Resources such as IndicNLPSuite~\cite{kakwani2020indicnlp} and Naamapadam~\cite{mhaske2023naamapadam} demonstrate the growing ecosystem of Indic NLP tools with which this framework can integrate.

\subsection{Limitations and Future Work}

Although the proposed framework achieves strong document-level performance, several limitations remain. The current prediction aggregation strategy simply retains the prediction from the most recently processed overlapping window. More sophisticated aggregation methods based on confidence scores, majority voting, or probability fusion may further improve boundary consistency.

Additionally, the evaluation was performed using synthetic long documents generated from the MahaNER test corpus~\cite{litake2022mahaner}. While these datasets enable controlled analysis of document length and contextual diversity, evaluation on naturally occurring long-document corpora would provide additional evidence of the framework's effectiveness in real-world scenarios, as suggested by prior work on long-range Marathi NER~\cite{deshmukh2024longrange}.

Future work will investigate adaptive window selection strategies, confidence-aware prediction aggregation, and evaluation on multilingual document-level NER benchmarks. Furthermore, integrating the proposed inference framework with emerging long-context transformer architectures~\cite{beltagy2020longformer,zaheer2020bigbird} may combine the advantages of efficient inference and enhanced contextual modeling.

\section{Conclusion}

This paper presented a training-free framework for extending sentence-level Named Entity Recognition (NER) models to document-level inference using an overlapping sliding window strategy. Instead of modifying the underlying transformer architecture~\cite{vaswani2017,devlin2019} or performing additional fine-tuning, the proposed approach partitions long documents into overlapping token windows, processes each window independently using a pretrained Marathi NER model~\cite{litake2022mahaner,joshi2022mahabert}, and aggregates the resulting predictions to generate coherent document-level annotations. This enables existing sentence-level models to process documents of arbitrary length while preserving compatibility with their original training configuration.

Extensive experiments were conducted on six document-level evaluation corpora derived from the MahaNER dataset~\cite{litake2022mahaner} using four different window--stride configurations. The results demonstrate that the proposed framework consistently maintains high Macro Precision, Macro Recall, and Macro F1-score~\cite{powers2011} across documents with varying lengths and contextual diversity. In particular, the smallest window configuration achieved the best overall performance, while larger window sizes resulted in only marginal reductions in recognition quality. These findings indicate that overlapping sliding window inference effectively preserves contextual continuity without requiring specialized long-context transformer architectures~\cite{beltagy2020longformer,zaheer2020bigbird} or additional annotated data.

The proposed framework provides a simple, computationally efficient, and easily deployable solution for document-level Named Entity Recognition in low-resource languages such as Marathi~\cite{lowresource}. Since the approach operates entirely during inference, it can be readily integrated with existing transformer-based NER models without retraining. Future work will focus on developing confidence-aware prediction aggregation strategies, adaptive window selection techniques, and evaluating the proposed framework on naturally occurring long-document corpora and multilingual document-level NER benchmarks, leveraging resources such as IndicNLPSuite~\cite{kakwani2020indicnlp} and Naamapadam~\cite{mhaske2023naamapadam}.

\section*{Acknowledgements}


This work was carried out under the mentorship of L3Cube, Pune. We would like to express our gratitude towards our mentor for his continuous support and encouragement. This work is a part of the L3Cube-MahaNLP project \cite{joshi2022l3cube_mahanlp}.


\bibliography{main}

\end{document}